# Analysis of Interpolation based Image In-painting Approaches

Mustafa ZOR[1], Erkan BOSTANCI[2], Mehmet Serdar GÜZEL[3*], Erinç KARATAŞ[4]

[1,2,3,4] Ankara University, Ankara,TR

**Corresponding author email[*]:** mguzel@ankara.edu.tr

Interpolation and internal painting are one of the basic approaches in image internal painting, which is used to eliminate undesirable parts that occur in digital images or to enhance faulty parts. This study was designed to compare the interpolation algorithms used in image in-painting in the literature. Errors and noise generated on the colour and grayscale formats of some of the commonly used standard images in the literature were corrected by using Cubic, Kriging, Radial based function and High dimensional model representation approaches and the results were compared using standard image comparison criteria, namely, PSNR (peak signal-to-noise ratio), SSIM (Structural SIMilarity), Mean Square Error (MSE). According to the results obtained from the study, the absolute superiority of the methods against each other was not observed. However, Kriging and RBF interpolation give better results both for numerical data and visual evaluation for image in-painting problems with large area losses.

## 1. Introduction

When analogue cameras were widely used, the photographs we kept in print were at risk of aging, fading, wear, and thus loss of information. With the advent of digital cameras, the development of computer and storage, and even cloud storage, our habit of storing photographs in print has evolved accordingly. However, this did not eliminate the risk of information loss of our photos. Errors or loss of information were observed in digital photographs during the acquisition and transmission of the photograph. Some of these may be related to the direct quality of the photograph, such as blur and noise, and in some cases, loss of information in certain areas of the picture. In addition, there may be errors such as the unintentional incorporation of undesirable objects into the photo frame during the photo shoot.

Some methods have been developed due to requirements such as eliminating the lack of information in photographs or eliminating unwanted areas. Bertalmio et al. (2000) developed in-painting terminology. In-painting is the art of modifying a picture or video in a way that cannot be easily detected by an ordinary observer and has become a major research area in image processing (Bugeau et al. 2010). The missing or undesirable portion of the image is completed by using intensity levels on adjacent pixels. However, the image in-painting will not be able to restore the original form of the missing part in the picture, it will only fill the missing or undesirable parts, close to the original (Amasidha et al. 2016).

The methods developed for in-painting can be grouped under three main headings.

- **Texture synthesis:** The basis of this approach is the self-similarity principle. It is based on the assumption that similar structures in a picture are often repeated (Bugeau et al. 2010).
- **Exemplar-based approach:** The missing region is filled with information from the known region at the patch level.
- **Partial differential equations and variation-based diffusion techniques:** The method of partial differential equations fills the regions to be uniformly propagated along the isophot directions from neighboring regions along the direction of isophot. The method, which gives good results for small areas, causes turbidity in larger areas. Variation methods address the problem in the form of finding the extremes of energy functions. However, these models only aim at dealing with non-textural in-painting. The difficulty of real in-painting problems is due to the rapid changes of the isophot and the roughness of the image functions (Chang and Chongxiu 2011).

In addition to this, methods have been developed that consider interiors as an interpolation problem and apply different interpolation techniques to complete the missing areas in the images (Guzel, 2015 and Mutlu et. al., 2014).

In this study, we aimed to evaluate the interpolation methods to handle the image in-painting process as an interpolation problem as the main contribution. State-of-the-art methods were tested on a generated dataset which was subject to both noise and corruption. These methods were then assessed based on their SSIM, PNSR and MSE in a quantitative evaluation. A qualitative evaluation was also performed on the results to ensure that the interpolation approach yielded visually appealing images (Bostanci, 2014 and Seref et. al, 2021).

The rest of the paper is structured as follows: Section 2 presents the current literature and background on the interpolation approaches employed in the paper. This section is followed by Section 3 where the dataset and the evaluation approach is elaborated. Section 4 presents both quantitative and qualitative results, finally the paper is concluded in Section 5.

## 2. Literature Review and Background
### 2.1 Cubic Interpolation
This interpolation is one of the common methods used to estimate unknown points. It generates results with smoother transitions and lower error rate than other interpolation techniques with polynomials. It is a commonly used method for filling lost pixels. When an interpolation of surface values is desired in a two-dimensional field, it can be formulated as follows.

On a unit square, if the surface values at the corners and the partial derivative values at these points are known, the convergence values of the points within the square can be obtained by polynomial (1)

$$p(x, y) = \sum_{i=0}^{3} \sum_{j=0}^{3} a_{ij} x^i y^j \qquad (1)$$

The known corner values of the surface and the partial derivative values $(f_x, f_y, f_{xy})$ in the x and y direction which can be obtained from the corner points. The polynomial is obtained by replacing the required equations with unknown coefficients for $a_{ij}$ which can be solved using matrix format to express the polynomial exactly.

### 2.2 Kriging Interpolation
Kriging is a geostatistics interpolation method that takes into account the distance and degree of variation between known points when estimating values at unknown points (Firas and Jassim 2013). This approach uses the values of the entire sample to calculate an unknown value. Kriging assumes that the distance or direction between sample points reflects a spatial correlation that can be used to explain variation on the surface. In kriging interpolation, a mathematical function is applied to all points of a specified number or a specified radius to determine the estimation value of each unknown point. The closer the point is, the higher value the weights have. Kriging is the most appropriate approach when there is a spatially related distance or directional trend in the data (spatial auto-correlation) and calculated as follows:

$$\hat{P}^* = \sum_{i=1}^{N} \lambda_i P_i \qquad (2)$$

where $N : k \times k$ (sample) is the total number of intact pixels, $\hat{P}^*$: pixel to be interpolated, $P_i : k \times k$ intact pixels in the block, $\lambda_i : k \times k$ weigths of the intact pixels in the block ($\sum_{i=1}^{N} \lambda_i = 1$)

. Kriging interpolation, chooses values of $\lambda_i$ in order to minimize the interpolation variance ($\sigma^2 = E\left[\left(P - \hat{P}^*\right)^2\right]$).

Two steps are required for Kriging interpolation:
1. Dependency rules: Constructing variograms and covariance functions to predict statistical dependence (spatial autocorrelation) based on the autocorrelation model (fitting a model).
The Variogram is a function of distance and direction that separates the two positions used to measure dependence. Variogram is defined as the variance of the difference between two variables at two different points and is calculated as:

$$2\gamma(h) = \frac{1}{n} \sum_{i=1}^{N} [P(x_i) - P(x_i + h)]^2 ; \quad P(x_i), P(x_i + h) : x_i, x_i + h \quad (3)$$

2. Estimation: Estimating unknown values.

The data is used twice to perform these two steps.

Jassim (2013) creates damaged images by using 4 different masks on 10 different grayscale images and uses Kriging interpolation to remove them. Sapkal and Kadbe (2016) also do in-painting using Kriging interpolation. In this study, the results are obtained by using the same masks on 5 different grayscale images.

Sapkal et al. (2016) used Kriging interpolation to remove several types of masks. In the study, the same dataset was used with a different set of masks.

Awati et al. (2017) conducted a study on troubleshooting colour images using modified Kriging interpolation. They separate colour images into RGB components and apply separate interpolation to each component. In practice, 3x3 matrices are used. The masks they use consist of vertical, horizontal and curved lines. They make separate trials for 1,2,3 and 4 pixel thicknesses as line thicknesses in each mask and evaluate the results.

**2.3 Radial Basis Functions**
Chang and Chongxiu (2011) define a mapping between the coordinates and colours of the image pixels, and implement an algorithm based on radial-based functions to generate the best approximation of this mapping in a given neighborhood. Radial-based functions are means of approximating a multivariable function as a linear combination of univariate functions. It is one of the methods that provides good results for the interpolation of scattered data. In order to increase the accuracy of their solutions and reduce the complexity of the algorithm, researchers create the pixel-by-pixel zoom function. For larger loss areas, interpolation of different overlapping coefficients is used.

Wang and Qin (2006) propose an algorithm for image in-painting based on compactly supported radial basis functions (CSRBF). The algorithm transforms the 2-D image in-painting problem from a 3-D point set into a surface reconstruction problem. First, a covered surface is constructed for approximation to the set of dots obtained from the damaged image using radial based functions. The values of the lost pixels are then calculated using this surface.

**2.4 High Dimensional Model Representation and Lagrange Interpolation**
Karaca and Tunga (2016), who consider image in-painting as an interpolation problem, have designed this problem by using the HDMR method and Lagrange interpolation, which allows a multivariate function to be expressed as the sum of multiple functions with less variables.

Normally grayscale images are represented as functions with two variables $f(x, y)$, $x$: *number of rows*, $y$: number of columns. Similarly, coloured images are represented by a function of three variables $f(x, y, z)$. In order to apply HDMR, the copy of the image itself is added as an additional dimension. In other words, a grayscale image, $f(x,y,n)$, $n=1,2$; and the colour image is expressed as $f(x,y,z,n)$, $z=1,2,3$, $n=1,2$ (RGB channels).

HDMR expansion for a multivariate function; fixed term is defined as the sum of functions of one variable, functions of two variables and others.

$$f(x_1, x_2, ..., x_N) = f_0 + \sum_{i_1=1}^{N} f_{i_1}(x_{i_1}) + \sum_{i_1,i_2=1}^{N} f_{i_1 i_2}(x_{i_1}, x_{i_2}) + ... + f_{12...N}(x_1, x_2, ..., x_N) \qquad (4)$$

In general, functions are represented up to univariate or bivariate functions, and the remainder is ignored as an approximation error. The function created for grayscale images can be fully represented when it is extended up to two-variable functions in the HDMR expansion (Altın and Tunga 2014). The researchers presented a representation by making up to three variable functions for in-painting with colour images and aimed to estimate the lost pixels by applying them with Lagrange interpolation.

In another study, Karaca and Tunga (2016) also studied the in-painting of a rectangular area using the same method. They tried 5x5, 10x10 and 20x20 pixel dimensions on different images for the in-painting area.

The algorithms used for in-painting in the literature have also been used for noise removal. Jassim (2013) tried the Kriging algorithm for salt & pepper noise reduction. For noise of varying intensity, it first *detects* noise using an 8x8 pixel filter on the image, and then applies Kriging interpolation to correct incorrect pixel values in this area.

## 3. Material and Method
### 3.1 Material
The internally stained images obtained from the methods used in the studies and their originals were presented with PSNR and SSIM criteria. In addition, although the masks used in the studies seem similar to each other, they have differences. This is a limitation for the exact comparison of algorithms. In order to make a full comparison of the algorithms proposed in this part of the study, we compared the results obtained by using the same masks and images and each of the methods.

256x256 grayscale and colour images were used to compare the algorithms. The images used were selected from the images commonly used in image processing research: Lena, Mandril, Peppers, Jetplane and House. The masks applied to the images are created as follows:
- Mask 1: simple curve, drawn with 4 pixel-thick pencil
- Mask 2: A non-condensed font consisting of several lines between 12-19 fonts,
- Mask 3: Intense font created with 12 font letters,
- Mask 4: Intense scratches of oblique, horizontal and vertical lines drawn with a 4-pixel-thick pencil,
- Mask 5: Frame created with a size of 40x40 pixels.

In addition, to test the noise reduction efficiency of interpolation methods, masks created with salt & pepper noise were applied. Their density levels are Noise 1: 10%, Noise 2: 30%, Noise 3: 50%, Noise 4: 70% and Noise 5: 90%, respectively. For instance, the presentation of the grayscale Lena image with the masks to be used is demonstrated in Table 1.

Table 1 Display of the masks used in the study on a sample image

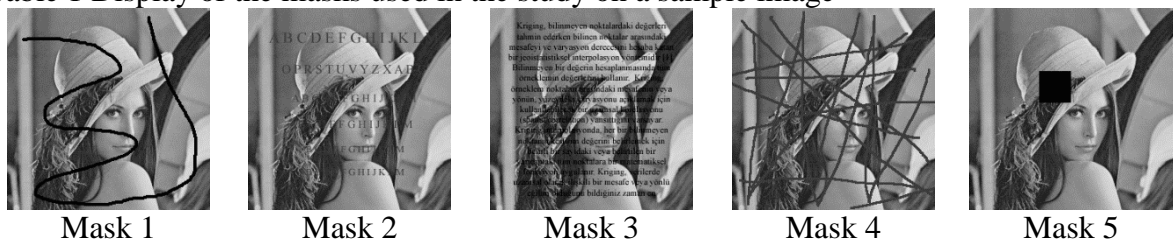

| Mask 1 | Mask 2 | Mask 3 | Mask 4 | Mask 5 |

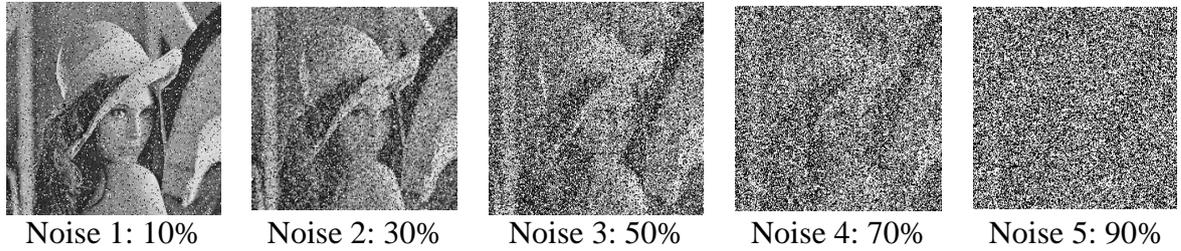

Noise 1: 10%  Noise 2: 30%  Noise 3: 50%  Noise 4: 70%  Noise 5: 90%

In all of the algorithms, corrected images were obtained by calculating only unknown pixel values and the obtained images were compared with the original image using PSNR (peak signal-to-noise ratio) and SSIM (Structural SIMilarity) criteria. In order to determine the unknown pixel values, the difference between the pixel values between the original image and the false image was taken and the non-0 pixel values were tried to be estimated by the algorithms. In the comparison of the results with the actual image, the mean square error (MSE), which is a part of the SSIM and PSNR criteria, is indicated.

### 3.2 Method

The evaluation presented in this study employs a number of various interpolation techniques, namely, Two-dimensional cubic interpolation, Kriging interpolation, Interpolation with radial based functions (RBF), Interpolation using High Dimensional Model Representation. These techniques are detailed in the following.

### 3.2.1 Two-dimensional cubic interpolation

For this method, x, y: row, column coordinate values of known pixels, v: pixel values, $x_q$, $y_q$: coordinate values of the desired pixels, cubic: to be used as input parameters for the interpolation. After the information of the two-dimensional image is put into the form that the function can use, the function is executed and the new pixel values are updated.

### 3.2.2 Kriging interpolation

For this method, kriging ($x_i$, $y_i$, $z_i$, x, y) and auxiliary functions presented by Schwanghart and Kuhn (2010) were used ($x_i$, $y_i$: row, column coordinate values of known pixels, $z_i$: pixel values, x, y: to be the value of the coordinate values of the desired pixels). 256x256 pixel images were used along with masks of size 16x16 pixels masks and noise densities defined in 8x8 pixels neighbourhood. The unknown values in each sub-image were calculated using Kriging function.

In this way, it is ensured that the unknown points in the sub-images are related to all pixel values in the sub-images. The 8x8 pixel size has enough data for noise reduction. However, the Kriging algorithm did not work for Noise5 with a 90% density (see Table 1). Therefore, only 16x16 pixels are used for Noise5. On the other hand, although the calculation method by sub-images is suitable for scattered errors, it will not be suitable for an internal painting problem as in Mask5. Therefore, for the case in Mask5, the image in-painting is reduced to a 90x90 pixel image size that will take the center of the lost frame area (40x40 pixels) and the function is run on this 90x90 pixel neighbourhood.

### 3.2.3 Interpolation with radial based functions (RBF)

The approach of (Foster 2009) was adopted in order to compute the RBF function in a similar fashion with the Kriging interpolation where subimages are used for the computation.

### 3.2.4 Interpolation using High Dimensional Model Representation

For this method, following the approach of Tunga and Koçanoğulları (2017) study; constant, one-variable and two-variable functions representing grayscale images; The fixed, one, two and three variable functions representing the coloured images were found and the missing areas on these functions were corrected by interpolation approach. In this section, spline for interpolation of one-dimensional functions and cubic interpolation for interpolation of two-dimensional functions were implemented.

These methods can be applied directly to grayscale images which can be expressed as two variable functions. Interpolation of coloured images were obtained by applying the above interpolations separately to the three layers of the images. At this point, there is no difference between the method applied to grayscale images. For the calculations, the same algorithms were applied to the red, green, blue (RGB) layers three times and then combined to produce corrected colour images.

**4. Results**

In-painting results were obtained using the various interpolation methods discussed above. All the results obtained for a sample are presented in Figures 1 (in grayscale) and 2 (in colour).

The numerical comparison results obtained from all images are summarized in Table 2 for grayscale and in Table 3 for colour images. The results of two-dimensional cubic interpolation, Kriging interpolation, RBF interpolation and YBMG interpolation are shown in four large blocks. Orange, blue, green intracellular staining was used to compare the outputs of the four methods. For example, in the PSNR values of 4 different methods used to remove Mask 1 in the colour Lena image, 2-dimensional cubic interpolation has the highest value. This cell was stained with orange. The highest SSIM values of the four methods were painted with blue, while the lowest MSE values of the four methods were painted with green. By means of this staining, it is better to observe which method gives better results in which situation.

In addition, the data of four different methods were compared with the one-way ANOVA test to determine whether there was a difference between the methods in terms of PSNR, MSE and SSIM. There was no difference between the methods in terms of PSNR, MSE and SSIM values in the evaluation (p = 0.997, p = 0.998, p = 0.986 for gray images, p = 0.974, p = 0.994, p = for gray images, respectively). 0988).

In addition, in order to compare the numerical results of the four methods, the difference between the maximum value and the smallest value of the PSNR and SSIM results obtained for each image and mask was attempted to be observed as a percentage change. These ratios are presented as Figures 3 and 4.

PSNR and MSE values are directly related to each other. As it is known, the MSE value is part of the PSNR criterion. The PSNR value is inversely proportional to the logarithm of the square root of MSE (5) Therefore, it is essential to have the lowest MSE when PSNR is highest.

$$PSNR = 20\log_{10}(255/\sqrt{MSE}) \qquad (5)$$

In addition, the formula of MSE (6) is built on differences in density levels between the two images.

$$MSE = \frac{1}{NM}\sum_{x=1}^{N}\sum_{y=1}^{M}\left[f(x,y) - g(x,y)\right]^2 \qquad (6)$$

In other words, the closer the pixel values in the same position in the two images are, the closer the MSE value is to 0. We understand that the smaller the MSE value and the higher the PSNR, the better the results. On the other hand, the SSIM criterion was formulated to measure luminance, contrast and structural correlation between the two images. The results obtained from functions comparing these three criteria separately are multiplied to find the SSIM value. Each function takes values in the range 0-1, as a result SSIM has values from 0 to 1. The SSIM value approaches to unity when the correlations are high, *i.e.* the images are similar. Although the PSNR and SSIM calculations do not seem to be alike, there are studies showing that these are analytically related (Horé and Ziou 2010). Therefore, SSIM and PSNR values are expected to be higher in a well obtained in-painting.

As a result of this evaluation; In Table 2 and Table 3, it will be natural for the cells stained with orange, blue and green to coexist. However, in some cases it is observed that this association is not preserved, but this is due to very small differences that can be ignored. Therefore, the outputs we obtained do not contradict theoretical knowledge.

For each image and method in Figure 3 and Figure 4, it was tried to observe the percentage change between the methods by the ratio of the highest and lowest obtained score differences to the lowest value. As can be observed here, the numerical comparison results for images obtained by in-painting are generally very close to each other. The difference in SSIM values does not exceed 5%. PSNR values are mostly within the 5% change band.

In addition, according to the statistics obtained from the one-way ANOVA test, it was confirmed that there was no difference between the methods in numerical evaluation scales. Therefore, it is observed that the methods do not have absolute superiority to each other in terms of in-painting results.

If the methods of application are taken into consideration, the in-painting problem in Mask 5 will need to be evaluated separately. In other masks, the problem of internal painting is scattered throughout the image, where it focuses on a specific area. Therefore, in Kriging and RBF interpolation methods, we focused on the 90x90 pixel area in order to keep the lost area in the middle. It is considered that the selected area would provide sufficient data in order to fill the missing area.

In fact, in the first trials, the whole image was tested using the Kriging and RBF functions at once for both the Mask 5 and the 4 other masks. However, the functions needed during the operation of the array and the required meshgrid structures that need to be produced. When the results obtained for Mask 5 are evaluated; it is observed that Kriging and RBF methods produce better outputs. On the other hand, this advantage will be determined in the evaluation made through observation. The results obtained in Mask 5 can be compared visually with the study of Karaca and Tunga (2016). It is observed that the results obtained for staining of the quadratic region are much larger than the MSE values.

| Corrupt Image | 2D Cubic Interpolation | Kriging Interpolation | RBF Interpolation | HDMR Interpolation |
|---|---|---|---|---|
| 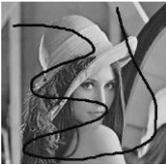 noisy image | 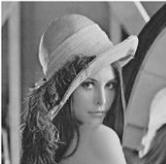 PSNR, SSIM, MSE 35.4708  0.98133  18.4501 | 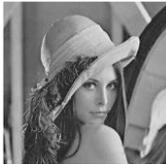 PSNR, SSIM, MSE 35.3839  0.98003  18.8231 | 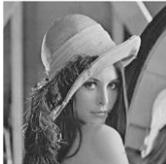 PSNR, SSIM, MSE 35.8737  0.98199  16.8157 | 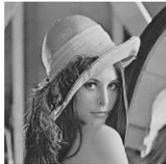 PSNR, SSIM, MSE 35.4708  0.98133  18.4501 |
| 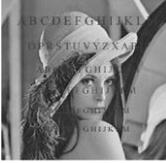 noisy image | 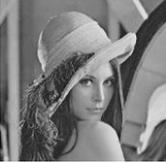 PSNR, SSIM, MSE 36.6386  0.98506  14.1002 | 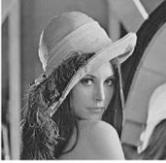 PSNR, SSIM, MSE 36.6824  0.98489  13.9586 | 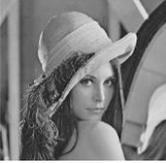 PSNR, SSIM, MSE 37.0019  0.98596  12.9686 | 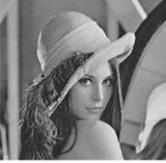 PSNR, SSIM, MSE 36.6386  0.98506  14.1002 |
| 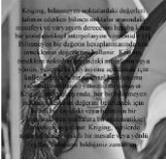 noisy image | 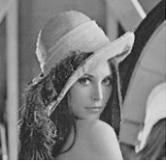 PSNR, SSIM, MSE 31.2984  0.94543  48.221 | 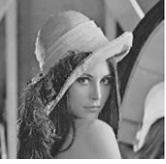 PSNR, SSIM, MSE 31.5  0.94765  46.0346 | 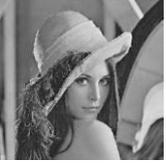 PSNR, SSIM, MSE 31.537  0.94927  45.6438 | 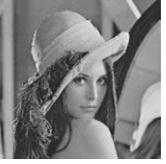 PSNR, SSIM, MSE 31.2984  0.94543  48.221 |
| 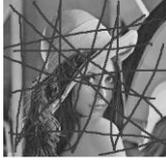 noisy image | 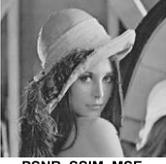 PSNR, SSIM, MSE 31.4843  0.95225  46.2008 | 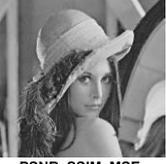 PSNR, SSIM, MSE 31.1682  0.94965  49.6887 | 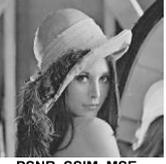 PSNR, SSIM, MSE 31.4559  0.95172  46.5044 | 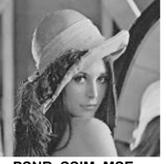 PSNR, SSIM, MSE 31.4843  0.95225  46.2008 |
| 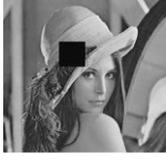 noisy image | 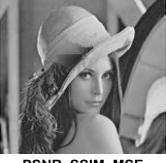 PSNR, SSIM, MSE 35.4137  0.98282  18.6943 | 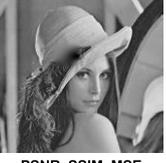 PSNR, SSIM, MSE 35.8538  0.98423  16.8927 | 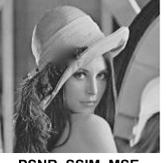 PSNR, SSIM, MSE 36.8819  0.98407  13.3318 | 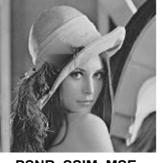 PSNR, SSIM, MSE 35.4137  0.98282  18.6943 |
| 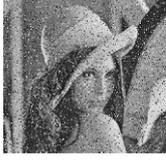 noisy image | 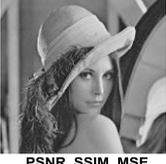 PSNR, SSIM, MSE 36.7836  0.9828  13.637 | 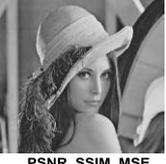 PSNR, SSIM, MSE 37.8719  0.98547  10.6143 | 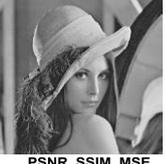 PSNR, SSIM, MSE 37.7299  0.98582  10.967 | 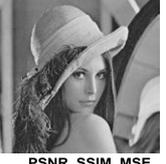 PSNR, SSIM, MSE 36.7836  0.9828  13.637 |
| 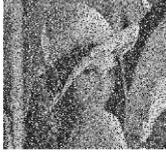 noisy image | 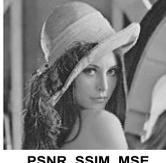 PSNR, SSIM, MSE 32.1876  0.94708  39.2938 | 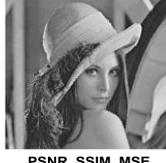 PSNR, SSIM, MSE 32.3014  0.9495  38.2768 | 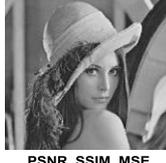 PSNR, SSIM, MSE 32.213  0.95041  39.0643 | 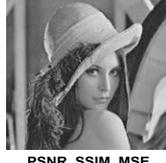 PSNR, SSIM, MSE 32.1876  0.94708  39.2938 |
| 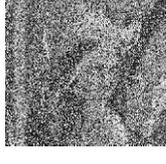 noisy image | 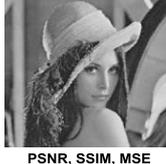 PSNR, SSIM, MSE 29.2525  0.90391  77.2376 | 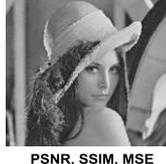 PSNR, SSIM, MSE 29.1843  0.90206  78.4611 | 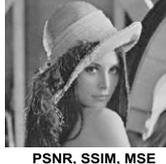 PSNR, SSIM, MSE 29.2039  0.90416  78.1066 | 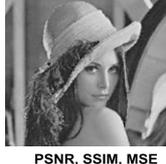 PSNR, SSIM, MSE 29.2525  0.90391  77.2376 |
| 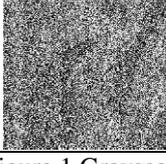 noisy image | 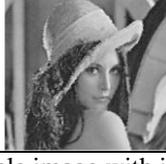 PSNR, SSIM, MSE 26.7749  0.84485  136.6437 | 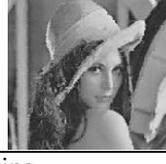 PSNR, SSIM, MSE 26.096  0.82037  159.7656 | 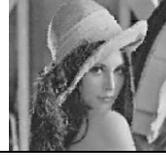 PSNR, SSIM, MSE 26.3119  0.82656  152.0166 | 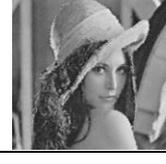 PSNR, SSIM, MSE 26.7749  0.84485  136.6437 |

Figure 1 Grayscale sample image with in-painting

| Corrupt Image | 2D Cubic Interpolation | Kriging Interpolation | RBF Interpolation | HDMR Interpolation |
|---|---|---|---|---|
| 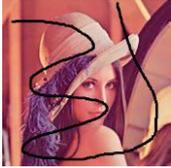 | PSNR, SSIM, MSE<br>37.3694 0.99657 11.9163<br>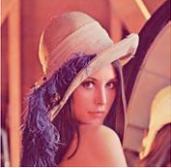 | PSNR, SSIM, MSE<br>36.3132 0.99598 15.1973<br>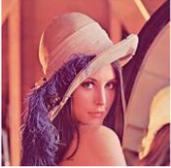 | PSNR, SSIM, MSE<br>36.1825 0.99586 15.6616<br>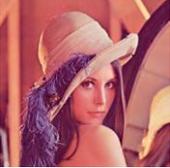 | PSNR, SSIM, MSE<br>37.3481 0.99656 11.9747<br>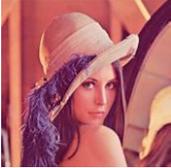 |
| 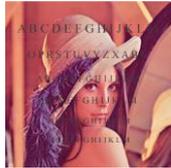 | PSNR, SSIM, MSE<br>39.0467 0.99774 8.0986<br>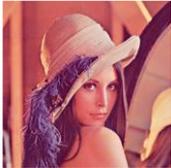 | PSNR, SSIM, MSE<br>37.8582 0.99718 10.6477<br>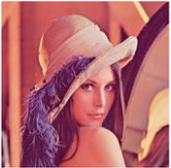 | PSNR, SSIM, MSE<br>37.5792 0.99682 11.3542<br>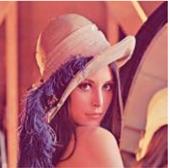 | PSNR, SSIM, MSE<br>39.0101 0.99773 8.1672<br>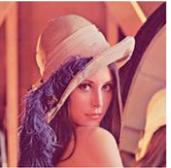 |
| 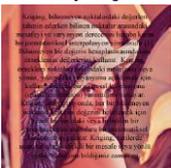 | PSNR, SSIM, MSE<br>33.0549 0.99038 32.1803<br>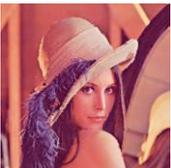 | PSNR, SSIM, MSE<br>32.5236 0.98954 36.3678<br>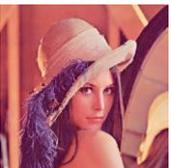 | PSNR, SSIM, MSE<br>32.4008 0.9892 37.4114<br>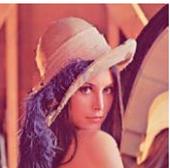 | PSNR, SSIM, MSE<br>33.0204 0.99032 32.4369<br>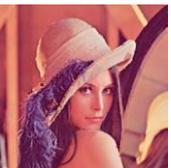 |
| 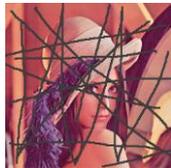 | PSNR, SSIM, MSE<br>32.8171 0.99044 33.9917<br>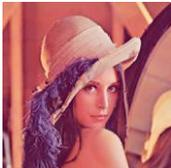 | PSNR, SSIM, MSE<br>31.3067 0.98766 48.1294<br>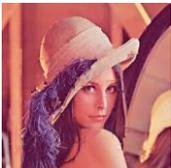 | PSNR, SSIM, MSE<br>31.2143 0.98711 49.1643<br>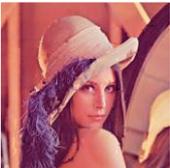 | PSNR, SSIM, MSE<br>32.8124 0.99044 34.0284<br>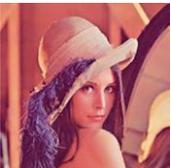 |
| 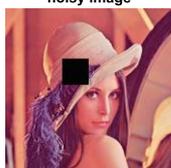 | PSNR, SSIM, MSE<br>36.1827 0.99591 15.6608<br>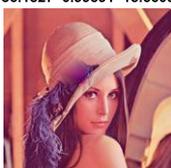 | PSNR, SSIM, MSE<br>36.3019 0.99621 15.2367<br>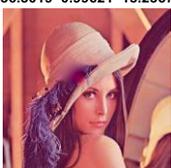 | PSNR, SSIM, MSE<br>36.7859 0.99637 13.63<br>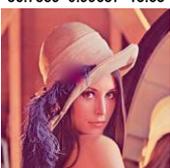 | PSNR, SSIM, MSE<br>36.1634 0.9959 15.7303<br>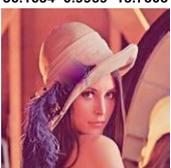 |
| 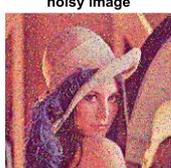 | PSNR, SSIM, MSE<br>39.9505 0.99786 6.577<br>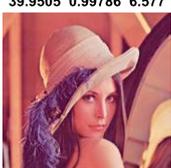 | PSNR, SSIM, MSE<br>41.0628 0.99835 5.091<br>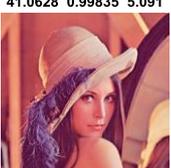 | PSNR, SSIM, MSE<br>40.443 0.99812 5.8719<br>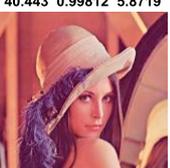 | PSNR, SSIM, MSE<br>39.9123 0.99785 6.6352<br>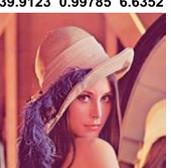 |
| 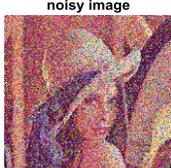 | PSNR, SSIM, MSE<br>34.9742 0.99332 20.685<br>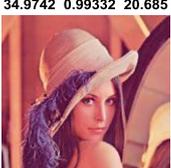 | PSNR, SSIM, MSE<br>34.9698 0.99342 20.7063<br>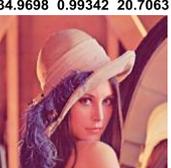 | PSNR, SSIM, MSE<br>34.5287 0.99277 22.9197<br>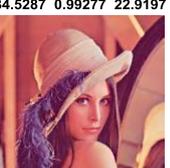 | PSNR, SSIM, MSE<br>34.9544 0.9933 20.7796<br>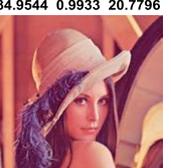 |
| 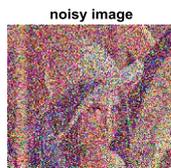 | PSNR, SSIM, MSE<br>31.5936 0.98569 45.0523<br>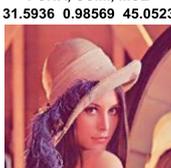 | PSNR, SSIM, MSE<br>31.2815 0.98493 48.409<br>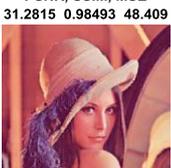 | PSNR, SSIM, MSE<br>31.1229 0.98414 50.2096<br>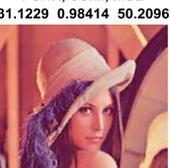 | PSNR, SSIM, MSE<br>31.5827 0.98567 45.1662<br>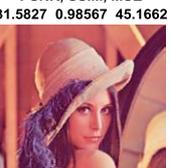 |
| 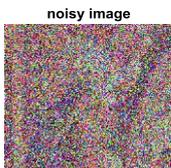 | PSNR, SSIM, MSE<br>28.6613 0.97254 88.501<br>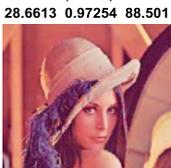 | PSNR, SSIM, MSE<br>27.9298 0.96815 104.7379<br>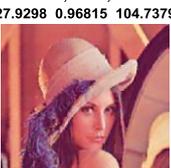 | PSNR, SSIM, MSE<br>28.0026 0.96827 102.9969<br>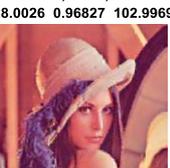 | PSNR, SSIM, MSE<br>28.659 0.97252 88.5482<br>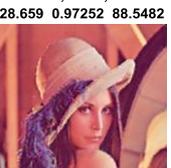 |

Figure 2 Colour sample image with in-painting

Table 2 Comparison of all images, masks and methods-Grayscale

| | | 2D Cubic Interpolation | | | Kriging Interpolation | | | RBF Interpolation | | | HDMR Interpolation | | |
|---|---|---|---|---|---|---|---|---|---|---|---|---|---|
| | | PSNR | SSIM | MSE | PSNR | SSIM | MSE | PSNR | SSIM | MSE | PSNR | SSIM | MSE |
| Lena | Mask1 | 35.471 | 0.98133 | 18.45 | 35.384 | 0.98003 | 18.823 | 35.874 | 0.98199 | 16.816 | 35.471 | 0.98133 | 18.45 |
| | Mask2 | 36.639 | 0.98506 | 14.1 | 36.682 | 0.98489 | 13.959 | 37.002 | 0.98596 | 12.969 | 36.639 | 0.98506 | 14.1 |
| | Mask3 | 31.298 | 0.94543 | 48.221 | 31.5 | 0.94765 | 46.035 | 31.537 | 0.94927 | 45.644 | 31.298 | 0.94543 | 48.221 |
| | Mask4 | 31.484 | 0.95225 | 46.201 | 31.168 | 0.94965 | 49.689 | 31.456 | 0.95172 | 46.504 | 31.484 | 0.95225 | 46.201 |
| | Mask5 | 35.414 | 0.98282 | 18.694 | 35.854 | 0.98423 | 16.893 | 36.882 | 0.98407 | 13.332 | 35.414 | 0.98282 | 18.694 |
| | Noise1 | 36.784 | 0.9828 | 13.637 | 37.872 | 0.98547 | 10.614 | 37.73 | 0.98582 | 10.967 | 36.784 | 0.9828 | 13.637 |
| | Noise2 | 32.188 | 0.94708 | 39.294 | 32.301 | 0.9495 | 38.277 | 32.213 | 0.95041 | 39.064 | 32.188 | 0.94708 | 39.294 |
| | Noise3 | 29.253 | 0.90391 | 77.238 | 29.184 | 0.90206 | 78.461 | 29.204 | 0.90416 | 78.107 | 29.253 | 0.90391 | 77.238 |
| | Noise4 | 26.775 | 0.84485 | 136.64 | 26.096 | 0.82037 | 159.77 | 26.312 | 0.82656 | 152.02 | 26.775 | 0.84485 | 136.64 |
| | Noise5 | 23.598 | 0.72678 | 283.94 | 23.019 | 0.69605 | 324.46 | 23.27 | 0.70052 | 306.23 | 23.598 | 0.72678 | 283.94 |
| House | Mask1 | 40.928 | 0.98931 | 5.2512 | 40.359 | 0.9893 | 5.9872 | 39.515 | 0.9883 | 7.2715 | 40.928 | 0.98931 | 5.2512 |
| | Mask2 | 43.898 | 0.99498 | 2.6504 | 43.479 | 0.99501 | 2.9186 | 42.548 | 0.99388 | 3.6162 | 43.898 | 0.99498 | 2.6504 |
| | Mask3 | 37.635 | 0.97906 | 11.211 | 37.862 | 0.97944 | 10.64 | 36.753 | 0.97635 | 13.733 | 37.635 | 0.97906 | 11.211 |
| | Mask4 | 35.842 | 0.97333 | 16.939 | 35.701 | 0.97317 | 17.497 | 34.682 | 0.96931 | 22.123 | 35.842 | 0.97333 | 16.939 |
| | Mask5 | 38.951 | 0.98915 | 8.2782 | 40.742 | 0.99042 | 5.4816 | 43.857 | 0.99163 | 2.6756 | 38.951 | 0.98915 | 8.2782 |
| | Noise1 | 47.819 | 0.99738 | 1.0744 | 47.855 | 0.99746 | 1.0657 | 45.775 | 0.99639 | 1.7201 | 47.819 | 0.99738 | 1.0744 |
| | Noise2 | 37.905 | 0.9874 | 10.535 | 39.997 | 0.98683 | 6.5063 | 38.573 | 0.98345 | 9.0317 | 37.905 | 0.9874 | 10.535 |
| | Noise3 | 36.898 | 0.97258 | 13.282 | 35.187 | 0.96611 | 19.697 | 34.445 | 0.96072 | 23.367 | 36.898 | 0.97258 | 13.282 |
| | Noise4 | 32.53 | 0.934 | 36.314 | 30.709 | 0.91736 | 55.226 | 30.35 | 0.90886 | 59.994 | 32.53 | 0.934 | 36.314 |
| | Noise5 | 25.212 | 0.82057 | 195.84 | 25.312 | 0.80812 | 191.38 | 25.289 | 0.80354 | 192.41 | 25.212 | 0.82057 | 195.84 |
| Peppers | Mask1 | 39.189 | 0.99088 | 7.8374 | 39.631 | 0.99074 | 7.0791 | 39.416 | 0.99066 | 7.4377 | 39.189 | 0.99088 | 7.8374 |
| | Mask2 | 42.063 | 0.99409 | 4.0435 | 42.656 | 0.99451 | 3.5274 | 41.299 | 0.99345 | 4.8212 | 42.063 | 0.99409 | 4.0435 |
| | Mask3 | 35.572 | 0.97824 | 18.025 | 35.484 | 0.97721 | 18.396 | 35.288 | 0.97772 | 19.243 | 35.572 | 0.97824 | 18.025 |
| | Mask4 | 32.549 | 0.96837 | 36.159 | 32.458 | 0.96667 | 36.926 | 31.916 | 0.96426 | 41.832 | 32.549 | 0.96837 | 36.159 |
| | Mask5 | 32.701 | 0.98492 | 34.916 | 36.59 | 0.99072 | 14.257 | 36.018 | 0.99031 | 16.264 | 32.701 | 0.98492 | 34.916 |
| | Noise1 | 42.558 | 0.99442 | 3.6084 | 42.956 | 0.99472 | 3.286 | 41.813 | 0.99402 | 4.2836 | 42.558 | 0.99442 | 3.6084 |
| | Noise2 | 36.44 | 0.9797 | 14.758 | 36.079 | 0.97823 | 16.037 | 35.309 | 0.97622 | 19.151 | 36.44 | 0.9797 | 14.758 |
| | Noise3 | 32.927 | 0.96034 | 33.146 | 32.572 | 0.9524 | 35.961 | 32.196 | 0.95017 | 39.22 | 32.927 | 0.96034 | 33.146 |
| | Noise4 | 28.73 | 0.91716 | 87.117 | 27.579 | 0.88713 | 113.55 | 27.886 | 0.88925 | 105.8 | 28.73 | 0.91716 | 87.117 |
| | Noise5 | 24.524 | 0.80588 | 229.46 | 24.03 | 0.77226 | 257.07 | 24.092 | 0.77667 | 253.46 | 24.524 | 0.80588 | 229.46 |
| Mandril | Mask1 | 33.607 | 0.9697 | 28.341 | 34.403 | 0.97272 | 23.591 | 34.308 | 0.97213 | 24.112 | 33.607 | 0.9697 | 28.341 |
| | Mask2 | 35.45 | 0.98171 | 18.537 | 35.744 | 0.98254 | 17.325 | 35.505 | 0.98199 | 18.307 | 35.45 | 0.98171 | 18.537 |
| | Mask3 | 29.277 | 0.92569 | 76.803 | 29.582 | 0.92779 | 71.59 | 29.385 | 0.92522 | 74.918 | 29.277 | 0.92569 | 76.803 |
| | Mask4 | 29.17 | 0.92005 | 78.714 | 29.423 | 0.92242 | 74.256 | 29.178 | 0.91994 | 78.572 | 29.17 | 0.92005 | 78.714 |
| | Mask5 | 37.325 | 0.98466 | 12.037 | 38.406 | 0.98865 | 9.3865 | 38.836 | 0.98899 | 8.5017 | 37.325 | 0.98466 | 12.037 |
| | Noise1 | 35.096 | 0.98013 | 20.113 | 35.138 | 0.98042 | 19.921 | 34.633 | 0.97827 | 22.374 | 35.096 | 0.98013 | 20.113 |
| | Noise2 | 29.509 | 0.92561 | 72.805 | 29.348 | 0.92245 | 75.555 | 29.118 | 0.91904 | 79.67 | 29.509 | 0.92561 | 72.805 |
| | Noise3 | 26.259 | 0.84231 | 153.87 | 26.297 | 0.83673 | 152.53 | 26.035 | 0.83151 | 162.03 | 26.259 | 0.84231 | 153.87 |
| | Noise4 | 23.592 | 0.71527 | 284.39 | 23.715 | 0.69501 | 276.4 | 23.381 | 0.69276 | 298.56 | 23.592 | 0.71527 | 284.39 |
| | Noise5 | 21.045 | 0.48207 | 511.14 | 21.325 | 0.46271 | 479.25 | 21.189 | 0.473 | 494.53 | 21.045 | 0.48207 | 511.14 |
| Jetplane | Mask1 | 35.763 | 0.98851 | 17.249 | 35.734 | 0.98738 | 17.363 | 35.419 | 0.98768 | 18.673 | 35.763 | 0.98851 | 17.249 |
| | Mask2 | 36.561 | 0.99187 | 14.355 | 36.541 | 0.99196 | 14.422 | 35.943 | 0.9909 | 16.548 | 36.561 | 0.99187 | 14.355 |
| | Mask3 | 31.514 | 0.96883 | 45.881 | 31.836 | 0.9692 | 42.605 | 31.48 | 0.96812 | 46.244 | 31.514 | 0.96883 | 45.881 |
| | Mask4 | 30.23 | 0.96264 | 61.668 | 30.528 | 0.96125 | 57.586 | 30.098 | 0.96007 | 63.578 | 30.23 | 0.96264 | 61.668 |
| | Mask5 | 34.77 | 0.99074 | 21.682 | 35.562 | 0.99056 | 18.067 | 35.175 | 0.99184 | 19.752 | 34.77 | 0.99074 | 21.682 |
| | Noise1 | 39.482 | 0.99356 | 7.3265 | 39.906 | 0.99405 | 6.6453 | 38.988 | 0.99309 | 8.208 | 39.482 | 0.99356 | 7.3265 |
| | Noise2 | 33.332 | 0.97603 | 30.188 | 33.393 | 0.97464 | 29.771 | 32.737 | 0.97209 | 34.625 | 33.332 | 0.97603 | 30.188 |
| | Noise3 | 29.752 | 0.94846 | 68.848 | 29.742 | 0.94317 | 69.002 | 29.321 | 0.93934 | 76.032 | 29.752 | 0.94846 | 68.848 |
| | Noise4 | 26.865 | 0.89929 | 133.82 | 25.929 | 0.87449 | 166.03 | 25.854 | 0.87406 | 168.91 | 26.865 | 0.89929 | 133.82 |
| | Noise5 | 22.372 | 0.7664 | 376.59 | 22.087 | 0.73366 | 402.12 | 22.301 | 0.74699 | 382.77 | 22.372 | 0.7664 | 376.59 |

Table 3 Comparison of all images, masks and methods-Colour

| | | 2D Cubic Interpolation | | | Kriging Interpolation | | | RBF Interpolation | | | HDMR Interpolation | | |
|---|---|---|---|---|---|---|---|---|---|---|---|---|---|
| | | PSNR | SSIM | MSE | PSNR | SSIM | MSE | PSNR | SSIM | MSE | PSNR | SSIM | MSE |
| Lena | Mask1 | 37.3690 | 0.9965 | 11.9160 | 36.3130 | 0.9959 | 15.1970 | 36.1820 | 0.9958 | 15.6620 | 37.3480 | 0.9965 | 11.9750 |
| | Mask2 | 39.0470 | 0.9977 | 8.0986 | 37.8580 | 0.9971 | 10.6480 | 37.5790 | 0.9968 | 11.3540 | 39.0100 | 0.9977 | 8.1672 |
| | Mask3 | 33.0550 | 0.9903 | 32.1800 | 32.5240 | 0.9895 | 36.3680 | 32.4010 | 0.9892 | 37.4110 | 33.0200 | 0.9903 | 32.4370 |
| | Mask4 | 32.8170 | 0.9904 | 33.9920 | 31.3070 | 0.9876 | 48.1290 | 31.2140 | 0.9871 | 49.1640 | 32.8120 | 0.9904 | 34.0280 |
| | Mask5 | 36.1827 | 0.9959 | 15.6608 | 36.3019 | 0.9962 | 15.2367 | 36.7859 | 0.9963 | 13.6300 | 36.1634 | 0.9959 | 15.7303 |
| | Noise1 | 39.9510 | 0.9978 | 6.5770 | 41.0630 | 0.9983 | 5.0910 | 40.4430 | 0.9981 | 5.8719 | 39.9120 | 0.9978 | 6.6352 |
| | Noise2 | 34.9740 | 0.9933 | 20.6850 | 34.9700 | 0.9934 | 20.7060 | 34.5290 | 0.9927 | 22.9200 | 34.9540 | 0.9933 | 20.7800 |
| | Noise3 | 31.5940 | 0.9856 | 45.0520 | 31.28200 | 0.9849 | 48.4090 | 31.1230 | 0.9841 | 50.2100 | 31.5830 | 0.9856 | 45.1660 |
| | Noise4 | 28.6610 | 0.9725 | 88.5010 | 27.9300 | 0.9681 | 104.7400 | 28.0030 | 0.9682 | 103.0000 | 28.6590 | 0.9725 | 88.5480 |
| | Noise5 | 24.5101 | 0.9364 | 230.1814 | 24.1843 | 0.9302 | 248.1129 | 24.2096 | 0.9303 | 246.6749 | 24.4822 | 0.9359 | 231.6658 |
| House | Mask1 | 40.4790 | 0.9959 | 5.8238 | 39.4160 | 0.9947 | 7.4390 | 39.0240 | 0.9945 | 8.1407 | 40.4790 | 0.9959 | 5.8238 |
| | Mask2 | 41.5140 | 0.9972 | 4.5883 | 41.4500 | 0.9970 | 4.6565 | 41.0230 | 0.9968 | 5.1379 | 41.5140 | 0.9972 | 4.5885 |
| | Mask3 | 35.6460 | 0.9892 | 17.7200 | 34.4480 | 0.9863 | 23.3500 | 34.0940 | 0.9858 | 25.3320 | 35.6460 | 0.9892 | 17.7200 |
| | Mask4 | 33.9880 | 0.9850 | 25.9600 | 32.7490 | 0.9813 | 34.5320 | 32.3050 | 0.9801 | 38.2500 | 33.9880 | 0.9850 | 25.9590 |
| | Mask5 | 40.0240 | 0.9977 | 6.4662 | 39.3290 | 0.9968 | 7.5886 | 41.6210 | 0.9986 | 4.4765 | 40.0240 | 0.9977 | 6.4662 |
| | Noise1 | 42.5520 | 0.9973 | 3.6128 | 41.9370 | 0.9972 | 4.1625 | 41.3550 | 0.9968 | 4.7599 | 42.7410 | 0.9973 | 3.4590 |
| | Noise2 | 35.5040 | 0.9889 | 18.3090 | 35.6730 | 0.9890 | 17.6100 | 35.3360 | 0.9880 | 19.0300 | 35.7270 | 0.9891 | 17.3920 |
| | Noise3 | 32.6610 | 0.9779 | 35.2370 | 32.2310 | 0.9763 | 38.9060 | 32.0950 | 0.9752 | 40.1400 | 33.0250 | 0.9783 | 32.4040 |
| | Noise4 | 29.3080 | 0.9565 | 76.2580 | 28.4140 | 0.9476 | 93.6950 | 28.5930 | 0.9482 | 89.9060 | 29.4980 | 0.9572 | 72.9990 |
| | Noise5 | 24.8730 | 0.8977 | 211.7300 | 24.5450 | 0.8880 | 228.3600 | 24.5710 | 0.8889 | 226.9800 | 25.0030 | 0.8994 | 205.5100 |
| Peppers | Mask1 | 36.7600 | 0.9964 | 13.7130 | 34.0110 | 0.9944 | 25.8200 | 35.1310 | 0.9955 | 19.9530 | 34.7740 | 0.9932 | 21.6610 |
| | Mask2 | 39.0040 | 0.9973 | 8.1792 | 38.1130 | 0.9970 | 10.0420 | 38.1140 | 0.9972 | 10.0400 | 36.1120 | 0.9942 | 15.9170 |
| | Mask3 | 32.7020 | 0.9898 | 34.9060 | 32.0700 | 0.9888 | 40.3730 | 32.0650 | 0.9892 | 40.4210 | 31.5940 | 0.9860 | 45.0530 |
| | Mask4 | 30.5920 | 0.9852 | 56.7340 | 28.9100 | 0.9800 | 83.5830 | 29.1750 | 0.9812 | 78.6360 | 29.7810 | 0.9812 | 68.3920 |
| | Mask5 | 32.510 | 0.9962 | 36.4840 | 35.3340 | 0.9976 | 19.0400 | 35.4160 | 0.9977 | 18.6830 | 31.7720 | 0.9934 | 43.2440 |
| | Noise1 | 37.0490 | 0.9960 | 12.8290 | 37.3210 | 0.9963 | 12.0500 | 37.3460 | 0.9964 | 11.9800 | 35.0560 | 0.9929 | 20.3010 |
| | Noise2 | 31.9010 | 0.9873 | 41.9780 | 31.9410 | 0.9876 | 41.5890 | 32.0020 | 0.9878 | 41.0100 | 30.7700 | 0.9830 | 54.4590 |
| | Noise3 | 29.2230 | 0.9769 | 77.7570 | 28.9520 | 0.9758 | 82.7650 | 29.0920 | 0.9765 | 80.1390 | 28.4840 | 0.9716 | 92.1980 |
| | Noise4 | 26.9580 | 0.9621 | 131.0100 | 26.0360 | 0.9550 | 161.9900 | 26.2500 | 0.9565 | 154.1800 | 26.0540 | 0.9523 | 161.3300 |
| | Noise5 | 23.1160 | 0.9209 | 317.3300 | 22.8300 | 0.9129 | 338.9300 | 22.8610 | 0.9130 | 336.5100 | 22.2150 | 0.8988 | 390.4300 |
| Mandril | Mask1 | 32.5990 | 0.9813 | 35.7410 | 32.1320 | 0.9806 | 39.8000 | 31.7410 | 0.9793 | 43.5520 | 32.5970 | 0.9813 | 35.7550 |
| | Mask2 | 34.1960 | 0.9884 | 24.7460 | 34.2940 | 0.9887 | 24.1930 | 34.0950 | 0.9882 | 25.3270 | 34.1920 | 0.9884 | 24.7660 |
| | Mask3 | 28.5525 | 0.9525 | 90.6530 | 28.4160 | 0.9498 | 93.6420 | 28.2680 | 0.9487 | 96.8830 | 28.5600 | 0.9525 | 90.5800 |
| | Mask4 | 28.4080 | 0.9535 | 93.8250 | 28.0750 | 0.9526 | 101.3000 | 27.5810 | 0.9496 | 113.4800 | 28.4070 | 0.9535 | 93.8400 |
| | Mask5 | 31.6180 | 0.9898 | 44.8000 | 30.8770 | 0.9922 | 53.1310 | 32.6760 | 0.9933 | 35.1160 | 31.6170 | 0.9898 | 44.8140 |
| | Noise1 | 33.9730 | 0.9854 | 26.0490 | 34.0290 | 0.9852 | 25.7150 | 33.7550 | 0.9844 | 27.3870 | 33.9720 | 0.9854 | 26.0570 |
| | Noise2 | 28.6800 | 0.9484 | 88.1250 | 28.7050 | 0.9475 | 87.6170 | 28.5010 | 0.9456 | 91.8380 | 28.6800 | 0.9484 | 88.1200 |
| | Noise3 | 25.6260 | 0.8935 | 178.0400 | 25.6400 | 0.8903 | 177.4400 | 25.5230 | 0.8893 | 182.2900 | 25.6230 | 0.89354 | 178.1300 |
| | Noise4 | 23.1580 | 0.8132 | 314.2800 | 23.0380 | 0.80268 | 323.0300 | 22.9140 | 0.8021 | 332.4000 | 23.1540 | 0.81314 | 314.5200 |
| | Noise5 | 20.4720 | 0.6666 | 583.3300 | 20.6280 | 0.66629 | 562.7500 | 20.5690 | 0.6665 | 570.4300 | 20.4700 | 0.66662 | 583.5400 |
| Jetplane | Mask1 | 37.1680 | 0.9923 | 12.4810 | 36.6330 | 0.9904 | 14.1190 | 35.7000 | 0.9897 | 17.5000 | 37.1600 | 0.99237 | 12.5040 |
| | Mask2 | 38.4190 | 0.9942 | 9.3570 | 37.6990 | 0.99307 | 11.0440 | 37.4830 | 0.9928 | 11.6100 | 38.4120 | 0.99428 | 9.3736 |
| | Mask3 | 32.2060 | 0.9772 | 39.1250 | 31.6920 | 0.97279 | 44.0430 | 31.4280 | 0.9729 | 46.8000 | 32.1880 | 0.97726 | 39.2910 |
| | Mask4 | 30.5990 | 0.9713 | 56.6480 | 29.3490 | 0.96121 | 75.5430 | 29.1340 | 0.9606 | 79.3740 | 30.5970 | 0.97132 | 56.6740 |
| | Mask5 | 35.4370 | 0.9909 | 18.5940 | 36.6620 | 0.99098 | 14.0250 | 36.6260 | 0.9922 | 14.1420 | 35.4320 | 0.99091 | 18.6170 |
| | Noise1 | 39.8120 | 0.9950 | 6.7904 | 39.7660 | 0.99492 | 6.8626 | 38.9670 | 0.9943 | 8.2479 | 39.7950 | 0.99509 | 6.8170 |
| | Noise2 | 33.7350 | 0.9817 | 27.5160 | 33.5000 | 0.97901 | 29.0470 | 33.0860 | 0.9781 | 31.9500 | 33.7190 | 0.98176 | 27.6190 |
| | Noise3 | 30.4710 | 0.9604 | 58.3400 | 29.6470 | 0.94927 | 70.5260 | 29.5550 | 0.9501 | 72.0480 | 30.4650 | 0.96044 | 58.4200 |
| | Noise4 | 27.0080 | 0.9208 | 129.5100 | 25.9770 | 0.8900 | 164.2200 | 25.8960 | 0.8922 | 167.2800 | 26.9900 | 0.9207 | 130.0300 |
| | Noise5 | 22.3250 | 0.8036 | 380.7200 | 22.4040 | 0.76772 | 373.8400 | 22.4520 | 0.7792 | 369.7000 | 22.3390 | 0.80356 | 379.4600 |

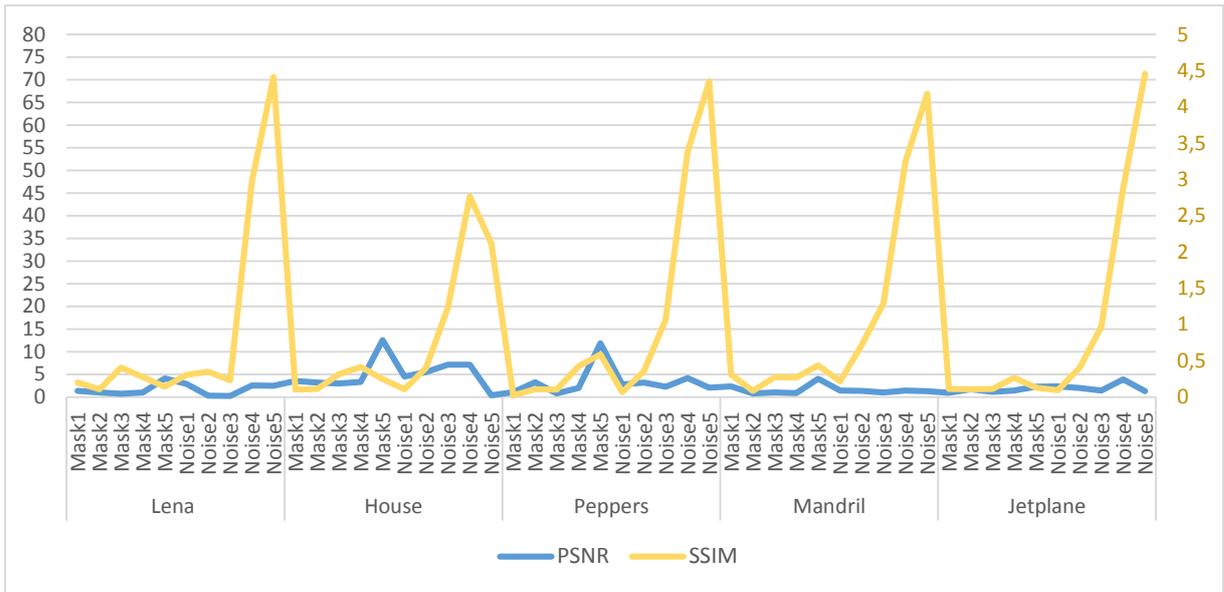

Figure 3 Percentage variation of the difference between the highest and lowest values of PSNR, SSIM scores obtained from grayscale images

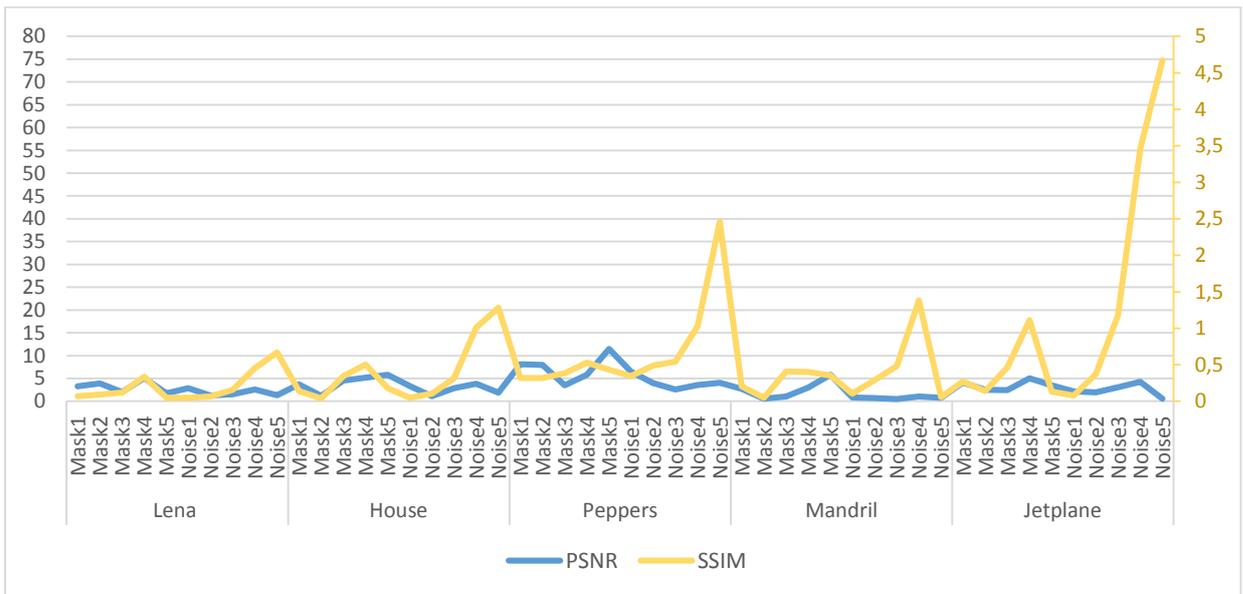

Figure 4 Percentage variation of the difference between the highest and lowest values of PSNR, SSIM scores obtained from colour images

According to the results, the square area is still clearly visible and filled with diagonal lines. In our results, the square area (for Kriging and RBF interpolation) was less prominent and no diagonal lines were observed in the faulty area. The difference between grayscale and colour images of the two studies may be inaccurate, but the interiors of in-painting characters can still be observed. Although relatively, the results obtained in our study can be more visually appealing.

## 5. Conclusion

The literature lacks a detailed evaluation on the use of interpolation methods for the in-painting problem. Accordingly, this study aimed to evaluate state-of-the-art approaches, namely, two dimensional cubic, Kriging, RBF and YBMG based Lagrange interpolation methods, which are recently studied in the literature.

A benchmark dataset was generated in order to perform a detailed evaluation of the in-painting algorithms. Several comprehensive experiments were conducted in order to make a fair assessment. According to the results obtained, it is observed that the methods do not have absolute superiority to each other in terms of in-painting results. However, Kriging and RBF interpolation produce better results both for numerical data and visual evaluation for image in-painting problems having large area losses. Studies on this subject have also yielded good results for large area in-painting, but Kriging and RBF interpolation outputs look better in terms of visual quality.